\documentclass{ifacconf}
\usepackage{graphicx}
\usepackage{float}
\usepackage{epsf}
\usepackage{amssymb}
\usepackage{fancyhdr}
\usepackage{subfigure}
\usepackage{latexsym}
\usepackage{amsmath}
\usepackage{natbib}

\newcommand{\diag} {\mbox{\rm diag}}

\newcommand{\virg}[1]{``#1''}

\begin{document}

\begin{frontmatter}

\title{Outlier robust system identification: a Bayesian kernel-based approach} 

\thanks[footnoteinfo]{The research leading to these results has received funding
from the Swedish Research Council under contract 621-2009-4017 and
the European Union Seventh Framework Programme
[FP7/2007-2013] under grant agreement no. 257462 HYCON2 Network of excellence, by the MIUR
FIRB project RBFR12M3AC - Learning meets time:
a new computational approach to learning in dynamic
systems}

\author[First]{Giulio Bottegal}
\author[Second]{Aleksandr Y. Aravkin}
\author[First]{H\r akan Hjalmarsson}
\author[Third]{Gianluigi Pillonetto}

\address[First]{ACCESS Linnaeus Centre, School of Electrical Engineering,
KTH Royal Institute of Technology, Stockholm, Sweden \\(e-mail: \{bottegal; hjalmars\}@kth.se)}
\address[Second]{IBM T.J. Watson Research Center, Yorktown Heights, NY, USA   (e-mail: saravkin@us.ibm.com)}
\address[Third]{Department of Information Engineering, University of Padova, Padova, Italy  (e-mail: giapi@dei.unipd.it)}


\begin{abstract}                          
In this paper, we propose an outlier-robust regularized kernel-based method for linear system identification.
The unknown impulse response is modeled as a zero-mean Gaussian process whose covariance (kernel)
is given by the recently proposed stable spline kernel, which encodes information on regularity and exponential stability.
To build robustness to outliers, we model the measurement noise as realizations of independent Laplacian random variables.
The identification problem is cast in a Bayesian framework, and solved by a new Markov Chain Monte Carlo (MCMC) scheme.
In particular, exploiting the representation of the Laplacian random variables as scale mixtures of Gaussians,
we design a Gibbs sampler which  quickly converges to the target distribution.
Numerical simulations show a substantial improvement in the accuracy of the estimates over state-of-the-art kernel-based methods.
\end{abstract}

\end{frontmatter}

\section{Introduction}
The classic approach to the problem of identifying a linear time-invariant system assumes that its transfer function belongs to a model class described by a small number
of parameters that determine important properties, such as zeros and poles positions, time constant, etc.
To identify the system, these parameters are estimated by minimizing a cost function related to the variance of the output prediction error.
This procedure, called prediction error method (PEM), is motivated by the fact that, when the number of available data tends to infinity, the parameter estimates are consistent and their variance attains the Cramer-Rao bound \citep{Ljung}, \citep{Soderstrom}. This  optimality result is guaranteed only when the \virg{true} model lies in the chosen model class. Clearly, in many situations choosing the appropriate model class may be an issue, and one should rely on model selection criteria such as AIC \citep{Akaike1974} or cross validation \citep{Ljung}. However, these criteria are consistent only asymptotically and may tend to overestimate the model order or provide poor predictive capability \citep{PitfallsCV12}.

Motivated by these issues, new identification paradigms have recently gained popularity.
Rather than positing a model class described by a small number of parameters and then estimating these,
newer methods try to estimate the entire impulse response.
In order to overcome the ill-posedness of this problem, these methods estimate \emph{hyperparameters} in order to regularize the identification process.
Hyperparameters can be seen as the counterpart of the parametric model order selection.
Kernel-based regularization methods are an important example of this kind of approach, and have had a long history in regression problems \citep{Poggio90}, \citep{Wahba1990}.
In the system identification framework, kernel-based methods have been introduced recently \citep{SS2010}, \citep{SS2011}.
The unknown impulse response is modeled as a realization of a Gaussian stochastic process,
whose covariance matrix belongs to the class of the so-called \emph{stable spline kernels} \citep{CDC2011P1}.
Introduced in \citep{SS2010}, kernels of this type have been proven to effectively model the behavior of the impulse response of stable systems \citep{ChenOL12},
exponential trends \citep{PillACC2010} and correlation functions \citep{bottegal2013regularized}.


In the kernel-based approach, the estimate of the impulse response is computed as the minimum variance Bayes estimate given the observed input/output data.
Recall that when the output is corrupted by white Gaussian noise, the impulse response and the output are jointly Gaussian.
However, if the white Gaussian noise assumption is violated, then the estimated impulse response may be poor.
In particular, this approach fails in the presence of outliers \citep{Aravkin2011tac}, \citep{Farahmand2011}; see the example below.

\subsection{A motivating example}
\begin{figure*}[!ht]
\begin{center}
\begin{tabular}{cc}
    {\includegraphics[width=7.5cm]{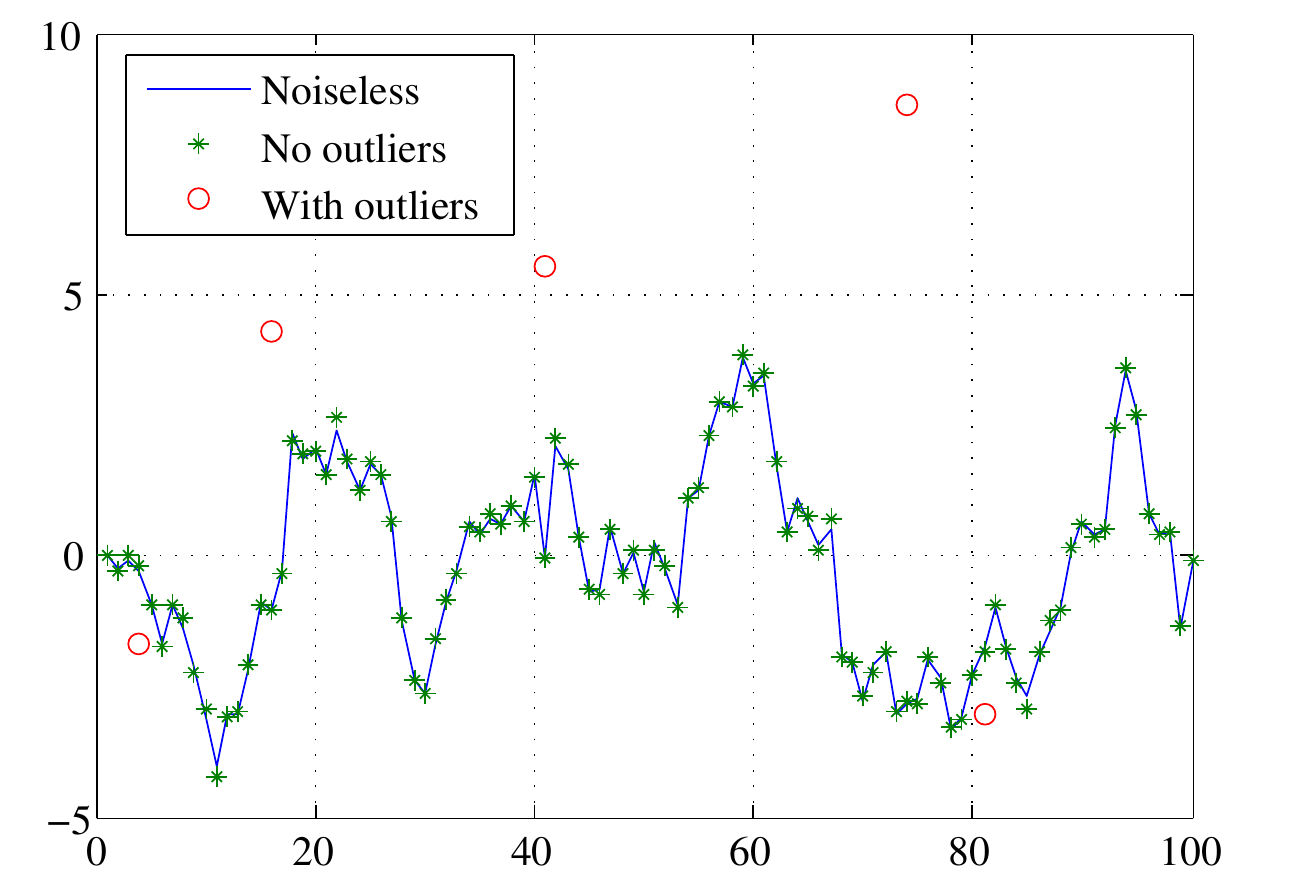}}
&
    {\includegraphics[width=7.5cm]{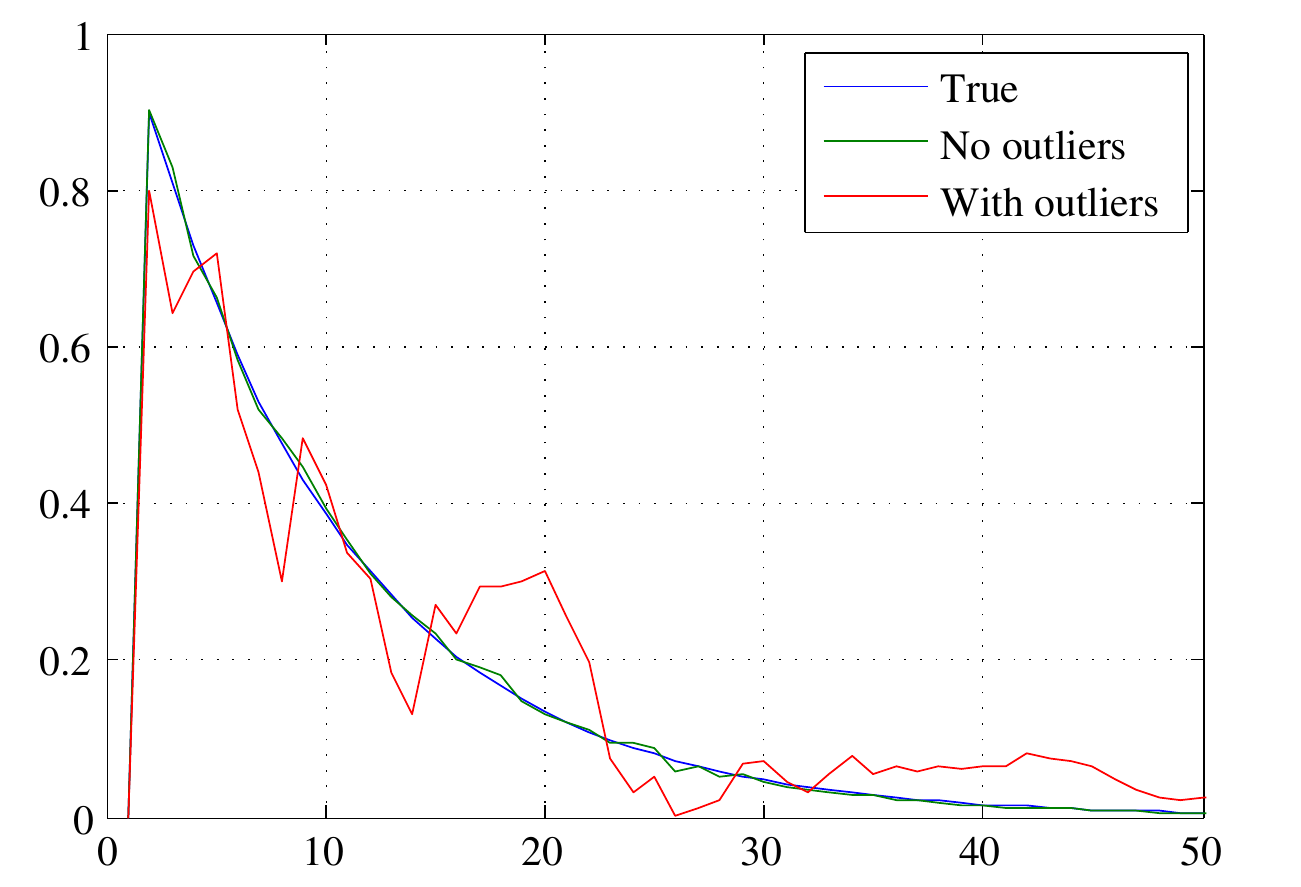}} \\
\end{tabular}
    \caption{\emph{Introductory example. Left panel: the noiseless output and the measured outputs in the no-outliers situation (measurements shown using green asterisks) and when outliers are present (shown using red circles). Right panel: the true impulse response and its estimate in the no-outliers situation and when outliers are present.}} \label{fig:example}
\end{center}
\end{figure*}

Suppose we want to estimate the impulse response of a linear system fed by white noise using the kernel-based method proposed in \citep{SS2010}. We consider two different situations, depicted in Figure \ref{fig:example}. In the first one, 100 samples of the output signal are measured with a low-variance Gaussian additive noise;
note that the estimated impulse response is very close to the truth.
In the second situation we introduce 5 outliers in the measured output,
obtaining a much poorer estimate of the same impulse response.
This suggests that outliers may have a devastating effect on the standard identification process that relies on the assumption of Gaussianity. 

\subsection{Statement of contribution and organization of the paper}
In this paper we introduce an outlier-robust system identification algorithm. 
We model the measurement noise as realizations of independent Laplacian random variables, which are better suited to modeling outliers because they have heavier
tails than the Gaussian distribution. 
Then, using stable spline kernels, we set a proper prior to the impulse response of the system, which allows us to to cast the problem into
a Bayesian framework and to solve it using Markov Chain Monte Carlo (MCMC) approach \citep{ParticleMCMC}. Note that MCMC-based approaches are  standard in system identification \citep{Ninness2010}, \citep{lindsten2012}.  A fundamental point of this work is exploiting the representation of Laplacian random variables as scale mixtures of Gaussians, that is, Gaussian variables whose variance has a prior exponential distribution. This representation allows us to design a Gibbs sampler \citep{Gilks},
which does not require any rejection criterion of the generated samples and quickly converges to the target distribution.
We evaluate the performance of the proposed algorithm using numerical simulations, and show
that in the presence of outliers, there is a substantial improvement of the accuracy of the estimated impulse response compared to the kernel-based method proposed in \citep{SS2010}.

The paper is organized as follows. In Section~\ref{sec:problem}, we formulate our system identification problem. In Section \ref{sec:bayesian} we cast
this problem in a Bayesian framework. In Section \ref{sec:algorithm}, we describe the proposed algorithm for impulse response estimation, and
test it using numerical simulations in Section \ref{sec:numerical}. Some conclusions end the paper.

\section{Problem statement}
\label{sec:problem}
\begin{figure}[!ht]
\begin{center}
{\includegraphics[width=6cm]{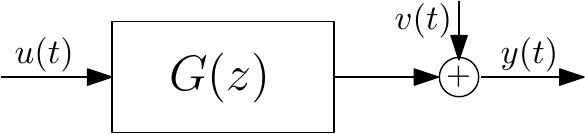}}
 \caption{\emph{Block scheme of the system identification scenario.}} \label{fig:block_scheme}
\end{center}
\end{figure}

We consider a SISO linear time-invariant discrete-time dynamic system (see Figure \ref{fig:block_scheme})
\begin{equation} \label{eq:sys1}
y(t) = G(z)u(t) + v(t) \,,
\end{equation}
where $G(z)$ is a strictly causal transfer function representing the dynamics of the system, driven by the input $u(t)$. The measurements of the output $y(t)$ are corrupted by the process $v(t)$, which is zero-mean white noise with variance $\sigma^2$. In the typical system identification framework,
the distribution of the noise samples is assumed to be Gaussian.
Here, instead, we consider a Laplacian probability density for the noise, i.e.
\begin{equation}
p(v(t)) = \frac{1}{\sqrt 2 \sigma} e^{- \frac{\sqrt 2 |v(t)|}{\sigma}} \,.
\end{equation}
We assume that $N$ samples of the input and output measurements are collected, and denote them by $u(1),\,\ldots,\,u(N)$, $y(1),\,\ldots,\,y(N)$.
Our system identification problem is to obtain an estimate of the impulse response $g(t)$ (or, equivalently, the transfer function) for $n$ time instants,
namely $\hat g(1),\,\ldots,\,\hat g(n)$. Recall that by choosing $n$ sufficiently large, these samples can be used to approximate $g(t)$ with arbitrary accuracy \citep{ljung1992}.

Introducing the vector notation
$$
y := \begin{bmatrix} y(1) \\ \vdots \\ y(N) \end{bmatrix} \,,\, g := \begin{bmatrix} 0 \\ g(1) \\ \vdots \\ g(n) \end{bmatrix} ,\, v := \begin{bmatrix} v(1) \\ \vdots \\ v(N) \end{bmatrix}
$$
$$
U = \begin{bmatrix} u(1) & 0  & & \ldots & 0 \\ u(2) & u(1) & 0 & \ldots & 0 \\ \vdots & \vdots & \ddots &\ddots & \\ u(N) & u(N-1)  &  \ldots & u(1) & 0 \end{bmatrix} \, \in \, \mathbb{R}^{N \times (n+1)}  \,,
$$
the input-output relation for the available samples can be written
\begin{equation} \label{eq:sys2}
y = Ug + v  \,,
\end{equation}
so that our estimation problem can be cast as a linear regression problem.
\section{A Bayesian framework} \label{sec:bayesian}
In this section we describe probabilistic models used for the quantities of interest in the problem.
\subsection{The stable spline kernel}
We first focus on setting a proper prior on $g$. Following a Gaussian regression approach \citep{Rasmussen}, we model $g$ as a zero-mean Gaussian random vector, i.e.
\begin{equation}\label{eq:model_g}
p(g) \sim \mathcal N (0,\,\lambda K_{\beta}) \,,
\end{equation}
where $K_\beta$ is a covariance matrix whose structure depend on the value of the parameter $\beta$ and $\lambda \geq 0$ is a scaling factor.
In this context, $K_\beta$ is usually called a {\it kernel}  and determines the properties of the realizations of $g$.
In this paper, we draw $K_\beta$ from the class of the stable spline kernels \citep{SS2010}, \citep{SS2011}.

There are two different types of stable spline kernels. The first one is defined by
\begin{equation} \label{eq:ssk1}
\{K_\beta\}_{i,j} := \beta^{ \max(i,j)} \quad,\, 0 \leq \beta \ <1 \,,
\end{equation}
and is known as \emph{first-order stable spline kernel} (or \emph{TC kernel} in \citep{ChenOL12}). The second type, known as \emph{second-order stable spline kernel}, is defined by
\begin{equation}\label{eq:ssk2}
\{K_\beta\}_{i,j} = \left[ \frac{\beta^{ (i+j)}
\beta^{\max(i,j)}}{2}-\frac{\beta^{3 \max(i,j)}}{6}\right], \quad 0 \leq \beta <1 \,.
\end{equation}
Compared to \eqref{eq:ssk1}, the latter type of stable spline kernel generates smoother impulse responses.

Both kernels \eqref{eq:ssk1} and \eqref{eq:ssk2} are parametrized by $\beta$, which regulates the decaying velocity of the generated impulse responses. Then, once the hyperparameters are fixed, the probability distribution of $g$ is
\begin{equation} \label{eq:prob_g}
p(g|\lambda,\,\beta) = \frac{1}{\sqrt{2 \pi \det (\lambda K_\beta)}} e^{-\frac{1}{2} g^T(\lambda K_\beta)^{-1}g} \,.
\end{equation}

Clearly, knowing the values of hyperparameters is of paramount importance to the design of an impulse response estimator.
The following result, drawn from \citep{MagniPAMI}, shows the marginal distribution of the inverse of the hyperparameter $\lambda$ given $g$ and $\beta$.

\begin{lem}\label{lem:lambda}
The posterior probability distribution of $\lambda^{-1}$ given $g$ and $\beta$ is
\begin{equation} \label{eq:Gamma_dist}
p(\lambda^{-1}|g,\,\beta) \sim \Gamma \left(\frac{n}{2}+1,\,g^T K_\beta^{-1} g \right)
\end{equation}
\end{lem}
\begin{rem}
To obtain the result of the above Lemma, we have implicitly set an improper prior on $\lambda$ with non-negative support. 
\end{rem}


\subsection{Modeling noise as a scale mixture of Gaussians} \label{sec:mixture}

The assumption on the noise distribution poses a challenge in expressing the conditional probability of $g$ given the input-output data, since it is non-Gaussian.
Here, we show how to deal with this problem. The key is to represent the noise samples $v(t)$ as a scale mixture of normals \citep{andrews1974}.
Specifically, denoting by $v_i$ the $i$-th entry of the noise vector $v$, for $i = 1,\,\ldots,\,N$, the pdf of $v_i$ can always be expressed as
\begin{equation}
p(v_i|\sigma^2) = \frac{1}{\sqrt 2 \sigma} e^{- \frac{\sqrt 2 |v_i|}{\sigma}} = \int_0^{+\infty}  \frac{1}{\sqrt{2 \pi \tau_i}} e^{- \frac{v_i^2}{2 \tau_i}} \frac{1}{\sigma^2} e^{- \frac{\tau_i}{\sigma^2}} d \tau_i \,.
\end{equation}
The above expression highlights the fact that each noise sample can be thought of as a realization of a Gaussian random variable, whose variance $\tau_i$ is in turn the realization of an exponential random variable, i.e.
\begin{equation} \label{eq:p_tau_i}
p(\tau_i|\sigma^2) =  \frac{1}{\sigma^2} e^{- \frac{\tau_i}{\sigma^2}}  \quad, \, \tau_i \geq 0 \,.
\end{equation}
Thus,
\begin{equation}
p(v_i|\tau_i,\,\sigma^2) = \frac{1}{\sqrt{2 \pi \tau_i}} e^{- \frac{v_i^2}{2 \tau_i}} \,
\end{equation}

The following result establishes a closed-form expression for the conditional probability density $p(\tau_i|v_i)$.
\begin{lem} \label{lem:GIG}
For any $i=1,\,\ldots,\,N$, the posterior of $\tau_i$ given $v_i$ is
\begin{equation} \label{eq:GIG_tau_i}
p(\tau_i|v_i,\,\sigma^2) \sim GIG\left(\frac{2}{\sigma^2},\,v_i^2,\,\frac{1}{2}\right) \,,
\end{equation}
that is generalized inverse Gaussian with parameters $(\frac{2}{\sigma^2},\,v_i^2,\,\frac{1}{2})$.
\end{lem}


Using the above result, we have that the posterior probability density of $\tau_i$ given $v_i$, $i = 1,\,\ldots,N$, is available in closed-form. The probability density \eqref{eq:GIG_tau_i} also depends on $\sigma^2$. Instead of establishing a prior for such a parameter, a consistent estimate of its value can be obtained with the following steps:
\begin{enumerate}
\item compute the least-squares estimate of $g$, i.e.
\begin{equation}
\hat g_{LS} = (U^T U)^{-1}U^T y \,,
\end{equation}
in order to obtain an unbiased estimate of $g$;
\item compute the empirical estimate of $\sigma_2$
\begin{equation} \label{eq:hat_sigma}
\hat \sigma^2 =  \frac{\left(y-U\hat g_{LS}  \right)^{T}\left(y-U\hat g_{LS} \right)}{N-n} \,.
\end{equation}
\end{enumerate}
In the following section, we shall assume that $\sigma^2$ is known.

\section{System identification under Gaussian and Laplacian noise assumptions} \label{sec:algorithm}

\subsection{The Gaussian noise case} \label{sec:gaussian_noise}
In this section, we make use of prior \eqref{eq:model_g} for modeling $g$,  assuming that the noise $v(t)$ is Gaussian. Then,  the joint distribution of the vectors $y$ and $g$, given values of $\lambda$, $\beta$ and $\sigma^2$, is jointly Gaussian, namely
\begin{equation} \label{eq:joint_Gaussian}
p\left(\left.\begin{bmatrix} y \\ g \end{bmatrix}\right|\lambda,\,\beta \right) \sim \mathcal N \left( \begin{bmatrix} 0\\0 \end{bmatrix} , \begin{bmatrix} \Sigma_y & \Sigma_{yg} \\ \Sigma_{gy} & \lambda K_\beta \end{bmatrix} \right)\,,
\end{equation}
where
\begin{equation} \label{eq:var_y}
\Sigma_y = \lambda U K_\beta U^T + \sigma^2 I_N
\end{equation}
and $\Sigma_{yg} = \Sigma_{gy}^T =  \lambda U K_\beta$. In this case, the minimum mean square error (MSE) estimation of $g$ is given by its Bayesian linear estimate, namely
\begin{equation} \label{eq:bayes_est1}
\hat E[g|y,\,\lambda,\,\beta] = \Sigma_{gy}\Sigma_y^{-1}y \,.
\end{equation}
The above equation depends on  unknown values of hyperparameters $\lambda$ and $\beta$. The estimate of such parameters, denoted $\hat \lambda$ and $\hat \beta$, can be performed by exploiting the Bayesian framework of the problem. More precisely, since $y$ and $g$ are jointly Gaussian, we can obtain $\hat \lambda$ and $\hat \beta$ by maximizing the marginal likelihood,  obtained by integrating out $g$ from the joint probability density of $(y,\,g)$.  Then we have
\begin{equation}\label{eq:marglik}
(\hat \lambda, \hat \beta) = \arg \min_{\lambda,\beta} \log \det (\Sigma_y) + y^T \Sigma_y^{-1} y \,.
\end{equation}
In this paper, we always use this approach to estimate $\beta$. Hence, below we shall consider such parameter to be known.

\subsection{The Laplacian noise case}
We now consider the proposed model, where $g$ has prior \eqref{eq:model_g} and the noise is modeled using the Laplacian distribution.
Then, the joint description of $y$ and $g$ given $\sigma^2$, $\lambda$ and $\beta$ does not admit a Gaussian distribution, since the vector $y$  is itself not Gaussian distributed. However, as shown in Section \ref{sec:mixture}, we can cast the problem in the Gaussian regression framework by introducing  variables $\tau_i,\,i=1,\,\ldots,\,N$.
In fact, it can be seen that, redefining $\Sigma_y$ as
\begin{equation}
\Sigma_y = \lambda U K_\beta U^T + D \,,\, D := \diag \{\tau_1,\ldots,\,\tau_N\}
\end{equation}
the joint posterior  of $y$ and $g$ given $\lambda, \beta, \sigma^2$ and all $\tau_i$ is again Gaussian:
\begin{equation} \label{eq:joint_Gaussian_tau}
p\left(\left.\begin{bmatrix} y \\ g \end{bmatrix}\right| \lambda,\,\{\tau_i\}_{i=1}^N\right) \sim \mathcal N \left( \begin{bmatrix} 0\\0 \end{bmatrix} , \begin{bmatrix} \Sigma_y & \Sigma_{yg} \\ \Sigma_{gy} & \lambda K_\beta \end{bmatrix} \right)\,,
\end{equation}
and the best estimator for $g$ is given by
\begin{equation} \label{eq:bayes_est2}
\hat E[g|y,\,\lambda,\,\{\tau_i\}_{i=1}^N] = \Sigma_{gy}\Sigma_y^{-1}y \,.
\end{equation}
Unfortunately, the above estimator requires the knowledge of the values of the $\tau_i$'s. In principle these parameters could be estimated by adopting a marginal likelihood function analogous to \eqref{eq:marglik}. However, the resulting minimization problems is extremely complicated and ill-posed, with a number of variables of the same order of the number of measurements and subject to multiple minima. Below, we describe our approach to solve the system identification problem.

\subsubsection{The proposed MCMC scheme}
The Bayesian approach to the problem permits to express the estimate of \eqref{eq:bayes_est2} as the following the integral
\begin{equation} \label{eq:bayes_est3}
\hat g = \int g\; p(g,\,\lambda,\,\{\tau_i\}_{i=1}^N|y)\; dg\; d\lambda\; \prod_{i=1}^{N} d \tau_i \,,
\end{equation}
which can be computed by Monte Carlo integration. In particular, it is sufficient to draw a large number of samples from the distribution $p(g,\,\lambda,\,\{\tau_i\}_{i=1}^N|y)$ and compute their average value, i.e.
\begin{equation}
\hat g = \lim_{M\rightarrow \infty} \frac{1}{M} \sum_{k=1}^M g^k \,,
\end{equation}
where the $g^k$ are used to denote these samples. Drawing samples from a distribution is a hard problem in general. However, when all the conditional probability densities of such a distribution are available in closed-form,
this can be done efficiently by employing a special case of the Metropolis Hastings sampler, namely the Gibbs sampler (see e.g. \citep{Gilks}).
The basic idea is that each conditional random variable is the state of a Markov chain; then, drawing samples from each conditional probability density iteratively,
we converge to the stationary state of this Markov chain and generate samples of the conditional distribution of interest. In our case, in view of \eqref{eq:bayes_est3}, we set
\begin{equation}
p(g,\,\lambda,\,\{\tau_i\}_{i=1}^N|y)
\end{equation}
as target probability density. Then, the conditional densities are as follows.
\begin{enumerate}
\item $p(\tau_i,\,|g,\,\lambda,\,\{\tau_j\}_{j=1,j\neq i}^N,\,y)$, $i = 1,\,\ldots,\,N$. Note that, for any $i = 1,\,\ldots,\,N$, $\tau_i$ is independent of $\lambda$, $\tau_j$ and $y_j$, $j \neq i$ and indeed it depends only on the observed value of the noise sample $v_i$. Then, recalling that $v_i = y_i - U_i g$, where $U_i$ denotes the $i$-th row of $U$, this conditional density has the form \eqref{eq:GIG_tau_i}, namely a generalized inverse Gaussian  with parameters $(\frac{2}{\sigma^2},\,(y_i - U_i g)^2,\,\frac{1}{2})$. Hence, it is available in closed-form.
\item $p(\lambda^{-1}|g,\,\{\tau_i\}_{i=1}^N,\,y)$. Once $g$ is given, $\lambda$ becomes independent of all the other variables (see Lemma \ref{lem:lambda}). Hence this conditional corresponds to the one stated in Lemma \ref{lem:lambda}, namely a Gamma distribution with parameters $(\frac{n}{2}+1,\,g^T K_\beta^{-1} g)$.
\item $p(g|\lambda,\,\{\tau_i\}_{i=1}^N,\,y)$. This probability density can be easily derived from \eqref{eq:joint_Gaussian_tau} and has a Gaussian distribution,
with mean $\lambda K_\beta U^T \Sigma_y^{-1}y$ and covariance $$\lambda K_\beta - \lambda^2 K_\beta U^T \Sigma_y^{-1} U K_\beta \,.$$
\end{enumerate}
Having established the above conditional probabilities, we need to specify the initial values for $g$ and $\lambda$, to be used as starting points in the iterative Gibbs sampler.
These are obtained by exploiting the estimation procedure proposed in Section \ref{sec:gaussian_noise} for the Gaussian noise case.

We now give our system identification algorithm.\\
\noindent\rule{\columnwidth}{0.3mm} \vspace{-.5cm}
\begin{algorithm}[ht!] \label{alg}
\textbf{Algorithm}: Outlier robust system identification \vspace{0.1cm}\\
Input: $\{y(t)\}_{t=1}^N,\,\{u(t)\}_{t=1}^N$ \vspace{0.1cm} \\
Output: $\{\hat{g}\}_{t=1}^n$
\begin{enumerate}
\item Initialization:
    \begin{enumerate}
        \item  Estimate $\sigma^2$ from \eqref{eq:hat_sigma} and $\beta$ from \eqref{eq:marglik}
        \item  Obtain $g^0$ from \eqref{eq:bayes_est1}  and $\lambda^0$ from \eqref{eq:marglik}
    \end{enumerate}
\item For $k =1$ to $M$:
    \begin{enumerate}
        \item  Draw the sample $\tau_i^k$, $i=1,\,\ldots,\,N$ from $$p(\tau_i,\,|g^{k-1},\,\lambda^{k-1},\,\{\tau_j^{k-1}\}_{j=1,j\neq i}^N,\,y)$$
        \item  Draw the sample $\lambda^k$ from
        $$p(\lambda^{-1}|\{\tau_i^{k}\}_{i=1}^N,\,g^{k-1},\,y)$$ 
        \item Draw the sample $g^k$ from
        $$p(g|\lambda^k,\,\{\tau_i^k\}_{i=1}^N,\,y)$$
    \end{enumerate}
\item Compute $\hat g = \frac{1}{M-M_0} \sum_{k=M_0}^M g^k$
\end{enumerate}
\end{algorithm}

\begin{figure}[!ht]
\begin{center}
{\includegraphics[width=\columnwidth]{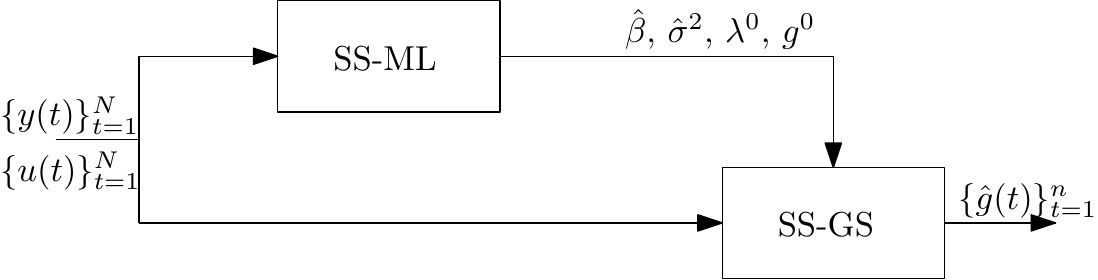}}
 \caption{\emph{Block scheme of the proposed algorithm. The label \emph{SS-ML} represents the marginal likelihood-based system identification method reviewed in Section \ref{sec:gaussian_noise}. The label \emph{SS-GS} indicates the Gibbs sampler step of the proposed method.}} \label{fig:alg_scheme}
\end{center}
\end{figure}

\vspace{-.5cm}
\noindent\rule{\columnwidth}{0.3mm}
In the  above algorithm, the parameters $M$ and $M_0$ are  introduced. $M$ the number of samples to be generated; clearly, large values of $M$ should guarantee more accurate estimates of $g$. $M_0$ is the number of initial samples drawn from the conditional of $g$ to be discarded. In fact, the conditionals from which those samples are drawn are to be considered as non-stationary, since the Gibbs sampler takes a certain number of iterations to converge to a stationary Markov chain.
\begin{rem}
The estimation procedure of $\beta$ is in a certain sense \virg{non-optimal}, since it is based on a different noise model.
However, we observe that the sensitivity of the estimator to the value of $\beta$ is relatively low, in the sense that a large interval of values of $\beta$ can model a given realization of $g$ efficiently (see Lemma 2 in \citep{bottegal2013regularized}). Models for $\beta$ will be introduced in future works.
\end{rem}
\begin{rem}
Notice that, differently from the empirical Bayes procedure described in the Gaussian noise case of Section \ref{sec:gaussian_noise}, the estimate $\hat g$ returned by the MCMC scheme designed for the Laplace noise case also accounts for the uncertainty related to $\lambda$ and $\tau_i$, $i=1,\,\ldots,\,N$.
\end{rem}
A block scheme representation of the proposed identification algorithm is shown in Figure \ref{fig:alg_scheme}. From this scheme,
it is clear that this algorithm can be seen as a refinement of the algorithm proposed in \citep{PillACC2010} and briefly described in Section \ref{sec:gaussian_noise}.

\section{Numerical experiments} \label{sec:numerical}
In this section, we report numerical results to illustrate the performance of the proposed algorithm.
We evaluate the proposed algorithm by means of 4 Monte Carlo
experiments of 100 runs each. At each run, a linear system is
randomly generated such that its transfer function $G(z)$ has 30 zeros and 30 poles. These poles are always within the
circle with center at the origin and radius 0.95 on the complex
plane. We consider an input-output delay equal to 1. In order to simulate the presence of outliers in the measurement process, the noise samples $v(t)$ are drawn from a mixture
two Gaussian of the form
\begin{equation}
v(t)\sim c_1 \mathcal N(0,\sigma^2) + c_2  \mathcal N(0,100\sigma^2) \,,
\end{equation}
with $c_1=0.7$ and $c_2=0.3$, so that outliers (observations with 100 times higher variance) are generated with probability 0.3.
The value of $\sigma^2$ was set to the variance of the noiseless output divided by 100.

Two different types of input signals are considered:
\begin{enumerate}
\item $u(t)$ is obtained by filtering a white noise sequence through a second-order low pass filter with random bandwidth (labeled as LP);
\item $u(t)$ is white noise (labeled as WN).
\end{enumerate}

At each Monte Carlo run, $N$ samples of
the input and output signals are generated; we consider two
different situations where the number of available samples is either
$N = 200$ or $N = 500$. In all the experiments, the parameter $n$
is set to $50$. Hence, there is a total of 4 different Monte Carlo experiments whose
features are summarized in Table 1. \\
{\small
\begin{center}
\begin{tabular}{|c|c|c|}
\hline Exp.$\#$ & Data set size ($N$) & Input type \\
\hline\hline $1$ & $200$ & LP \\
\hline $2$ & $500$ & LP \\
\hline $3$ & $200$ & WN \\
\hline $4$ & $500$ & WN \\
\hline
\end{tabular}
\\ \vspace{0.1cm} \emph{\small Table 1: Features of the 4 Monte Carlo experiments.}
\end{center}
}

Two different algorithms are tested; their performances
are evaluated at any run by computing the fitting score, i.e.
\begin{equation}
FIT_i (\%) = 100\left(1-\frac{\|g_i - \hat g_i \|_2}{\|g_i\|_2}\right) \,,
\end{equation}
where $g_i$ and $\hat g_i$ represent, respectively, the true
and the estimated impulse responses (truncated at the $n$-th sample) obtained at the $i$-th run.
The estimators tested are specified below.

\begin{itemize}
\item {\it{SS-ML}}: This is the nonparametric kernel-based identification method proposed in \citep{PillACC2010},
revisited in \citep{ChenOL12} and briefly described in Section \ref{sec:gaussian_noise}.
The impulse response is modeled as in \eqref{eq:model_g} and the hyperparameters $\lambda$ and $\beta$ are estimated by using a marginal likelihood maximization approach.
Note that this estimator does not attempt to model the presence of outliers.
\item {\it{SS-GS}}: This is the approach proposed in this paper, where a Gibbs sampler is employed for computing \eqref{eq:bayes_est3}. The parameter $M$, denoting the number of samples generated from each conditional probability density, is set to $1500$. The first $M_0 = 500$ generated samples are discarded. The validity of the choice of $M$ and $M_0$ is checked  by assessing that quantiles 0.25, 0.5, 0.75 are estimated with good precision \citep{Raftery1996}. The initial values of $g$ and $\lambda$ and the estimated values of $\beta$ and $\sigma^2$ are drawn from the {\it{SS-ML}} Algorithm.
\end{itemize}

\begin{figure*}[!ht]
\begin{center}
\begin{tabular}{cc}
    {\includegraphics[width=7.5cm]{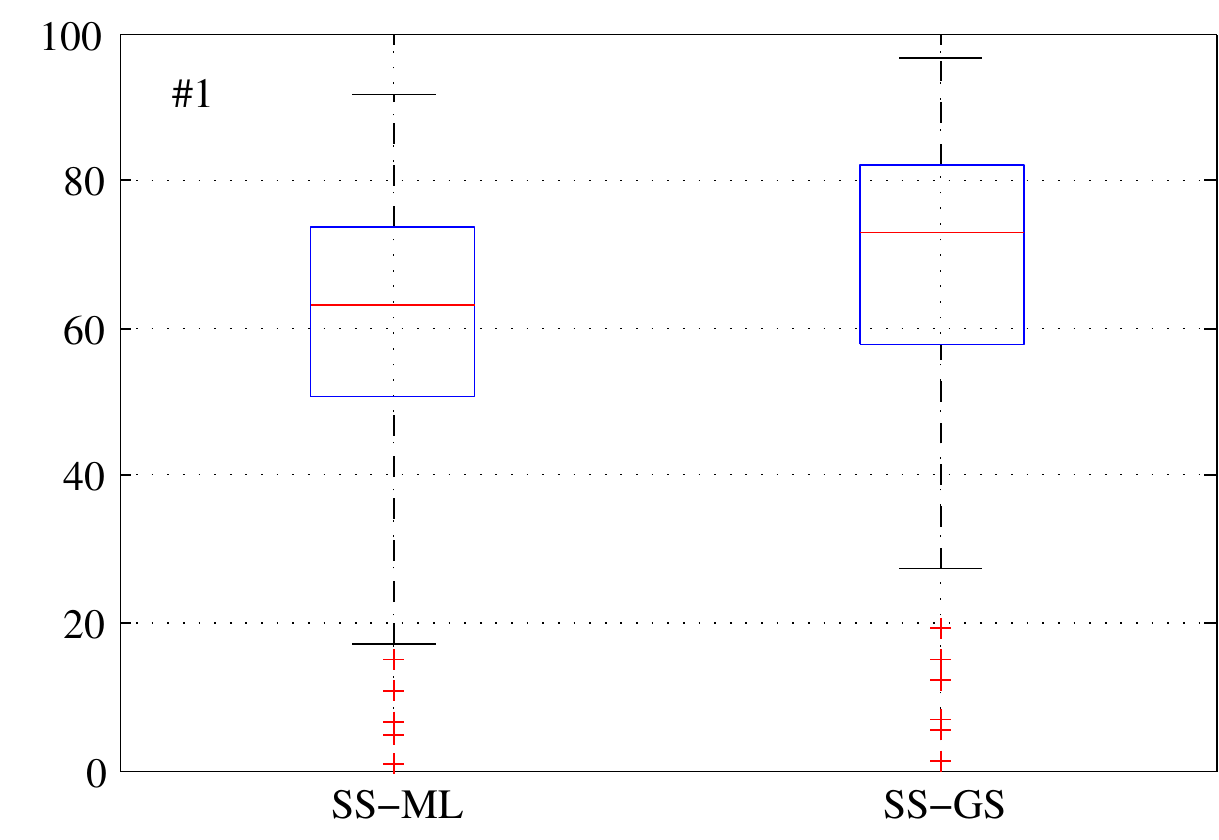}}
&
    {\includegraphics[width=7.5cm]{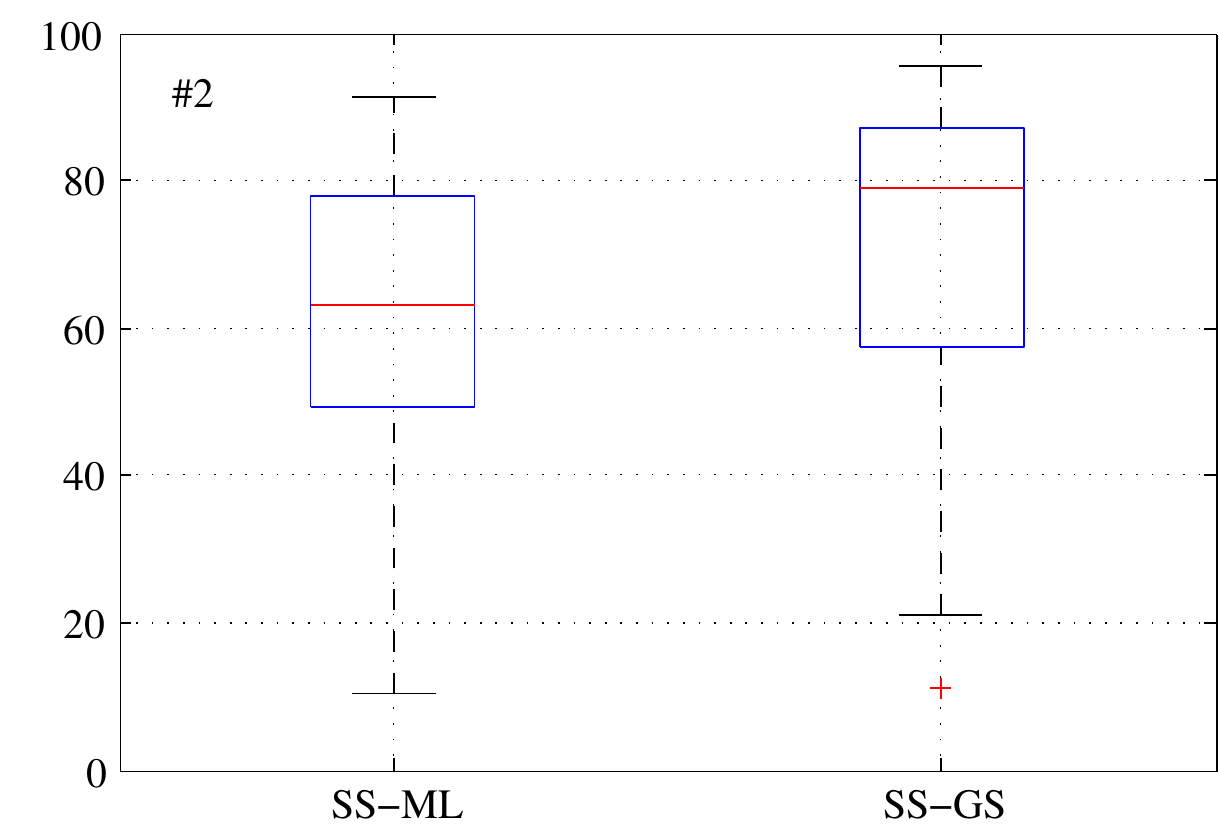}} \\
        {\includegraphics[width=7.5cm]{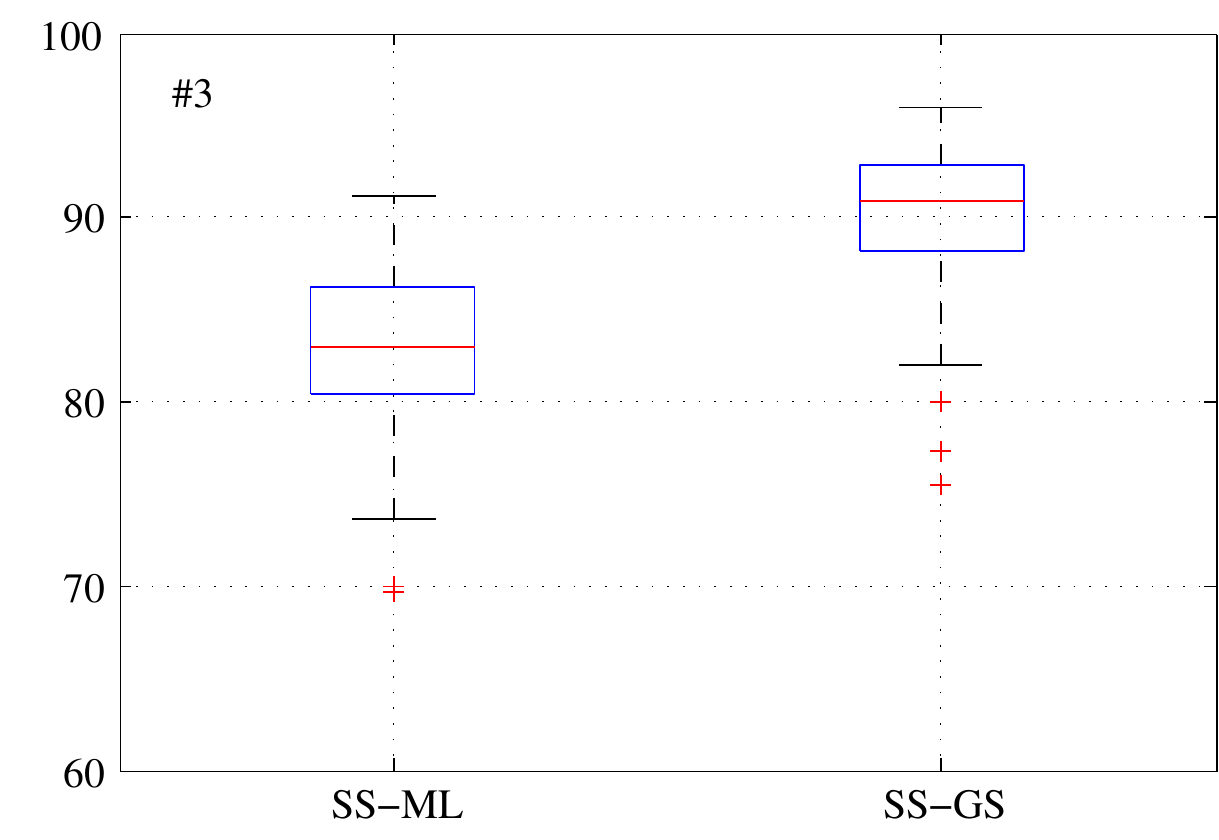}}
&
    {\includegraphics[width=7.5cm]{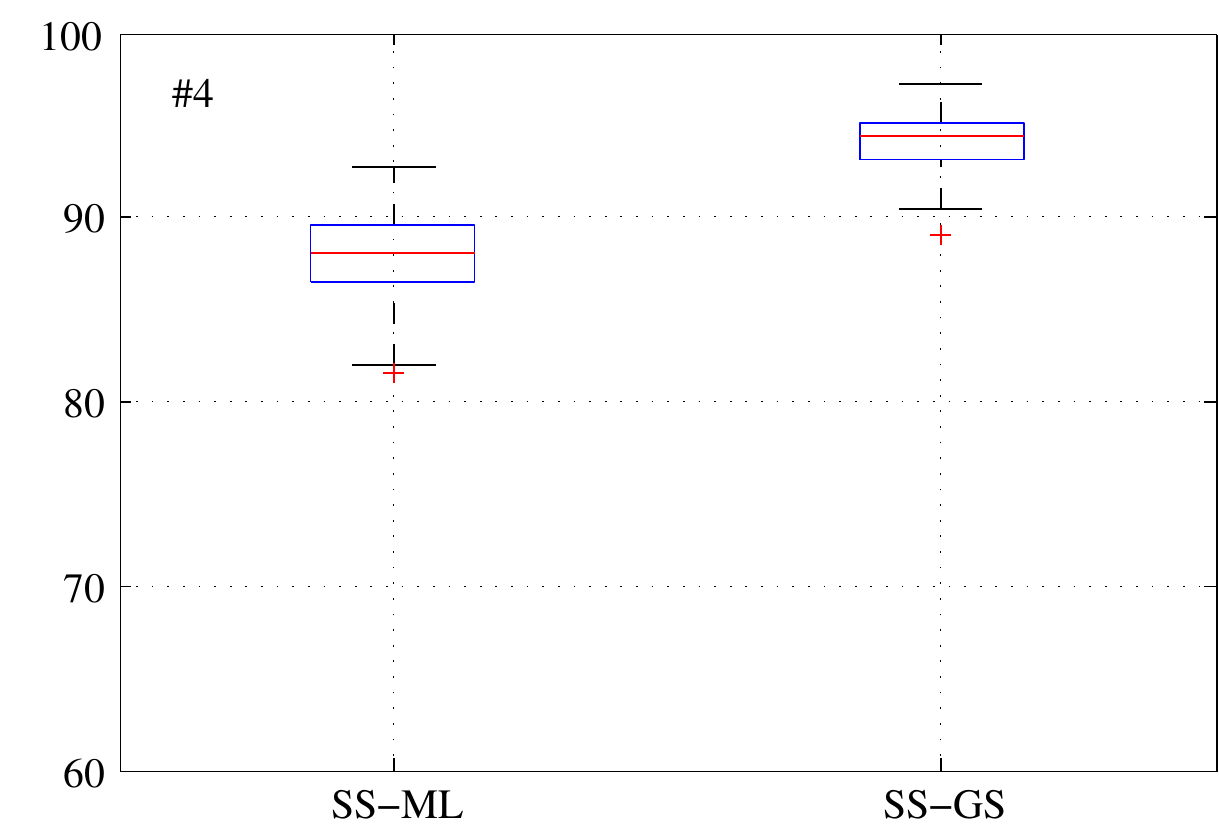}} \\
\end{tabular}
    \caption{\emph{Box plots of the fitting scores when measurements are corrupted by outliers.  The description of the experiments is summarized in Table 1.}} \label{exp_res1}
\end{center}
\end{figure*}

Figure \ref{exp_res1} shows the box plots of the 100 reconstruction
errors obtained by each estimator after the 4 Monte Carlo
experiments. The proposed method offers a substantial improvement of the fitting score in the example scenario.
This is particularly visible in the case of white noise, where the fitting score is above $90 \%$.
When the input is a low-pass signal, one can see that sometimes the performance of the estimators are not so satisfactory.
This happens when a high-pass transfer function  is fed with a short-band input, a combination that is known to give rise to ill-posed problems \citep{Bertero1}.

\subsection{An example with no outliers}
\begin{figure}[!ht]
\begin{center}
{\includegraphics[width=7.5cm]{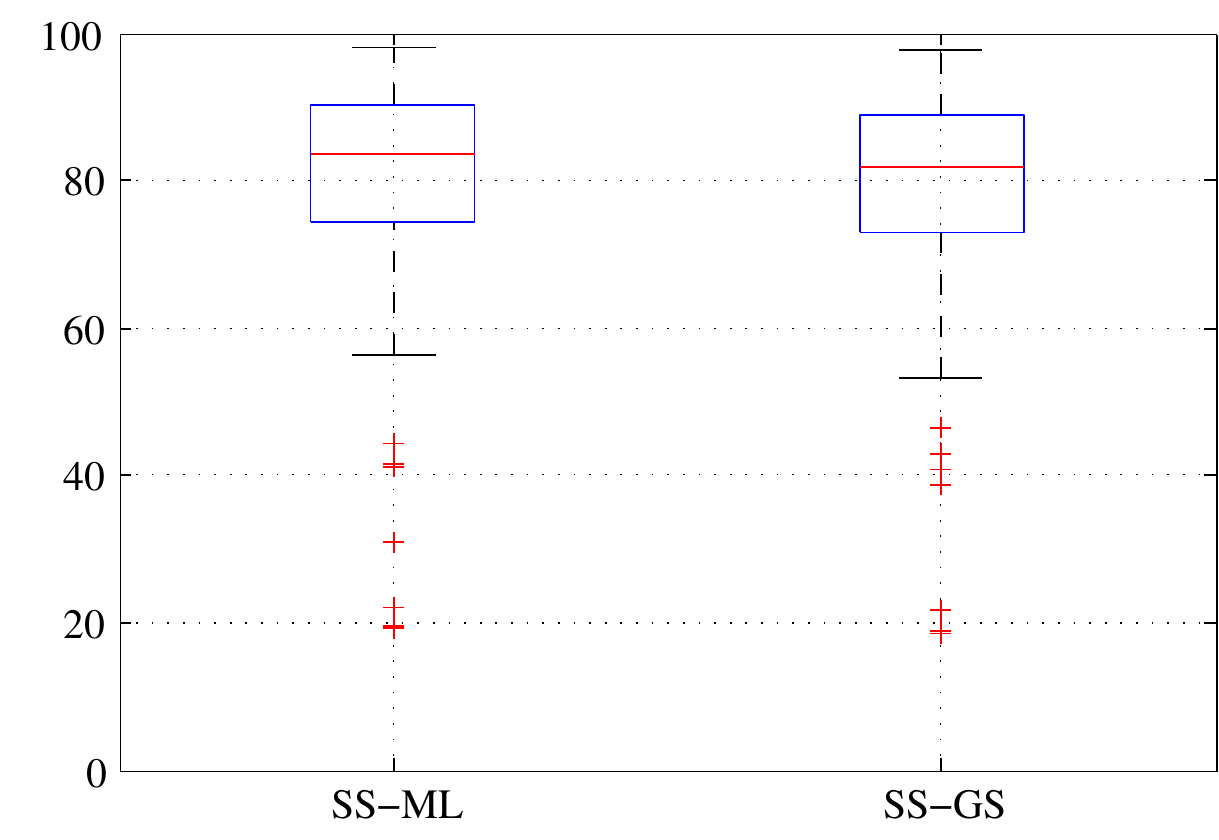}}
 \caption{\emph{Box plot of the fitting score when no outliers are simulated.}} \label{exp_res2}
\end{center}
\end{figure}

In order to complete our analysis, we also test our algorithm in the same framework as above, but setting $c_1=1$ and $c_2=0$, that is, generating errors
from a Gaussian noise model with no outliers. We use $N=500$, and generate inputs by filtering white noise through random second order low-pass filters.
The boxplots of Figure \ref{exp_res2} show the comparison between SS-ML and SS-GS over 100 Monte Carlo runs.
The performance of the proposed algorithm is comparable to the performance of the SS-ML Algorithm, with a slight degradation in the fitting score
due to the modeling of the noise, which, in the proposed estimator, is Laplacian instead of Gaussian.

\section{Conclusions}
In this paper, we have proposed a novel identification scheme for estimating impulse responses of linear system when the measurements are corrupted by outliers. We have shown that, modeling the measurement noise as Laplacian random variables, we can model our problem using a mixture of Gaussian random variables.
The mixture coefficients can be estimated by adopting a MCMC scheme which exploits closed-form expressions of conditional probabilities for the parameters of interest.
The performance of the proposed algorithm gives a substantial improvement over the state-of-the-art algorithm, which does not use outlier-robust noise modeling.

\section*{Appendix}
\subsection*{Proof of Lemma \ref{lem:GIG}}

We have
\begin{align}
p(\tau_i|v_i) & = \frac{p(v_i|\tau_i)p(\tau_i)}{p(v_i)} \\
                & = \frac{1}{\sqrt{2\pi \tau_i}} e^{-\frac{v_i^2}{2 \tau_i}} \frac{1}{\sigma^2} e^{-\frac{\tau_i}{\sigma^2}} \frac{\sqrt 2 \sigma}{e^{-\frac{\sqrt 2 |v_i|}{\sigma^2}}} \\
                & = \frac{ \frac{1}{\sigma}}{\sqrt{2} \sqrt{\frac{\pi}{2}} e^{-\frac{\sqrt 2 |v_i|}{\sigma^2}} }\tau_i^{-\frac{1}{2}} e^{-\frac{1}{2}\left(\frac{2\tau_i}{\sigma^2} + \frac{v_i^2}{\tau_i} \right)} \\
                & = \frac{\left(\frac{2}{v_i^2\sigma^2} \right)^{\frac{1}{4}} \left(\frac{2 v_i^2}{\sigma^2} \right)^{\frac{1}{4}}}{2 \sqrt{\frac{\pi}{2}} e^{-\frac{\sqrt 2 |v_i|}{\sigma^2}}} \tau_i^{-\frac{1}{2}} e^{-\frac{1}{2}\left(\frac{2\tau_i}{\sigma^2} + \frac{v_i^2}{\tau_i}  \right)}  \,.
\end{align}
Now, recalling that,  when $p = 1/2$ the modified Bessel function of second kind $K_p(z)$ has the form
\begin{equation}
K_{1/2}(z) = \sqrt{\frac{\pi}{2}} e^{-z} z^{\frac{1}{2}} \,,
\end{equation}
one can easily observe that, defining
\begin{equation}
a := \frac{2}{\sigma^2} \quad,\quad b := v_i^2 \quad,\quad p := \frac{1}{2} \,,
\end{equation}
one has
\begin{equation}
\sqrt{\frac{\pi}{2}} e^{-\frac{\sqrt 2 |v_i|}{\sigma^2}} \sqrt{\left(\frac{2 v_i^2}{\sigma^2} \right)^{-\frac{1}{2}}} = K_{1/2}(\sqrt{ab})
\end{equation}
and
\begin{equation}
\left(\frac{2}{v_i^2\sigma^2} \right)^{\frac{1}{4}} = \left(\frac{a}{b} \right)^{\frac{p}{2}} \,,
\end{equation}
so that
\begin{equation}
p(\tau_i|v_i) = \frac{\left(\frac{a}{b} \right)^{\frac{p}{2}}}{2  K_{p}(\sqrt{ab})} \tau^{p-1} e^{-\frac{1}{2}\left(a\tau_i + \frac{b}{\tau_i}  \right)} \,,
\end{equation}
that is $p(\tau_i|v_i) \sim GIG(a,\,b,\,p)$.
\hfill{$\Box$}

\bibliographystyle{plain}
\bibliography{biblio}

\end{document}